\documentclass[10pt]{article} 

\newif\ifpreprint
\preprinttrue

\ifpreprint
\usepackage[preprint]{tmlr}
\else
\usepackage{tmlr}
\fi

\usepackage{amsmath,amsfonts,bm}

\def\eqref#1{equation~\ref{#1}}

\def\1{\bm{1}}

\DeclareMathAlphabet{\mathsfit}{\encodingdefault}{\sfdefault}{m}{sl}
\SetMathAlphabet{\mathsfit}{bold}{\encodingdefault}{\sfdefault}{bx}{n}

\usepackage{hyperref}
\usepackage{url}
\usepackage{mathtools}
\usepackage{graphicx}
\usepackage{adjustbox}
\usepackage{amsmath}
\usepackage{amssymb}
\usepackage{centernot}
\usepackage{multirow}
\usepackage{multicol}
\usepackage{color}
\usepackage{mdframed}
\usepackage{subfigure}
\usepackage{caption}
\usepackage{dsfont}
\usepackage{enumerate}
\usepackage{enumitem}
\usepackage{empheq}
\usepackage{amsfonts}
\usepackage{totcount}

\usepackage{tabularx}
\usepackage{booktabs}
\usepackage{multirow}
\usepackage{colortbl}
\usepackage{makecell}
\usepackage{xcolor}
\usepackage{changepage}
\usepackage{stfloats}
\usepackage{soul}
\usepackage[ruled,vlined]{algorithm2e}
\usepackage{algpseudocode}
\usepackage[title]{appendix}
\usepackage{pdfpages}

\usepackage{bm}
\usepackage{amsmath,amssymb,amsfonts}
\usepackage{textcomp}
\usepackage{bookmark}
\usepackage{float}
\usepackage{tabularx}
\usepackage{mathtools}
\usepackage{array}
\usepackage{rotating}
\usepackage[skins]{tcolorbox}
\usepackage{tikz}
\usepackage{pifont}
\usepackage{lineno}

\newtheorem{definition}{Definition}%

\title{To Use AI as Dice of Possibilities with Timing Computation}

\author{Jia Li$^{* \ddagger}$ (corresponding author),  Vipin Kumar$^{\ddagger}$, Rui Zhang$^{*}$\\
\addr $^{*}$ Department of Surgery, University of Minnesota, Minneapolis, United States\\
$^{\ddagger}$ Department of Computer Science \& Engineering, University of Minnesota, Minneapolis, United States\\
\email  jiaxx213@umn.edu, kumar001@umn.edu, ruizhang@umn.edu
}

\def\month{Apr}  
\def\year{2023} 
\def\openreview{\url{https://openreview.net/forum?id=XXXX}} 

\definecolor{thegrey}{RGB}{192,192,192}
\definecolor{orchid}{RGB}{204,153,255}
\definecolor{myblue}{RGB}{0,115,207}
\definecolor{mygreen}{RGB}{68,134,25}
\definecolor{mypurple}{RGB}{177,35,200}

\begin{document}

\maketitle

\ifpreprint
\vspace{-4.5mm}
\fi

\begin{abstract}
\vspace{-3mm}

\ifpreprint
\vspace{-1mm}
\fi

The dominant noun-based modeling paradigm, grounded in probability theory and committed to pre-specified noun entities as primitive modeling units, is insufficient as a \emph{grammar of thought}:
It leaves \emph{timing} outside the computational scope, precluding any adequate representation of the future as an open space of possibilities.

This paper addresses three conceptual gaps absent from the existing literature: (1) possibility space --- a framework admitting multiple possible timelines for the same event; (2) timing computation --- the treatment of timing as a computable rather than observed dimension; and (3) causal factum --- the maximal causal efficacy recovered by reasoning backward from possible futures, rather than assumed in advance. 
Together, these definitions dissolve the confounding problem inherent to noun-based causal inference and provide the foundation for a spontaneously growing causal-reasoning world model. 

As proof of concept, we instantiate the framework and apply it to longitudinal EHR data from 3,276 breast cancer patients, demonstrating for the first time, to our knowledge, automatic trajectory discovery and counterfactual timing deduction (i.e., a What-If Machine) in a purely data-driven manner.
\ifpreprint
\cite{li2026capturetimingattentioneventsclinical}.
\else
\cite{anonymous2026companion}.
\fi

\end{abstract}

\ifpreprint
\vspace{-3mm}
\fi

Contemporary modeling paradigms, including those underlying the now-dominant large language models (LLMs) and meta-learning approaches, are grounded in probability theory, and thereby inherit a common set of metaphysical commitments: every modelable object must be a well-defined, stable individual entity --- a token, a feature, a data point --- a ``noun,'' in effect; and the relationships between such entities are encoded as functional mappings.

However, this noun-centric ontology proves insufficient as a \textit{grammar of thought}, that is, it cannot adequately represent the full range of conceptual problems that genuinely arise in human reasoning.

Consider a clinical example. Statistically, patients who have completed only one cycle of chemotherapy far outnumber those who have completed three --- simply because fewer patients reach that stage. Yet a doctor does not treat the first group as more important just because it is larger.
From a physician's standpoint, both situations represent real, co-existing possibilities for any given patient: the first toward imminent recovery or death; the second toward improved long-term survival but with elevated recurrence risk. 
No clinician would conclude that the first trajectory is therefore more clinically significant, simply because of statistical predominance, but would instead reason across all possible futures and pursue the best one for each individual patient.

Our thought is, at its core, oriented toward choosing among possible futures --- from which we can identify two keywords of the \emph{grammar of thought}: choice and future. 
These are also the two fundamental demands that thought places on any language, including the computational language of machine learning.
Yet in real-world settings, the variables that could influence an outcome are open-ended and, in principle, inexhaustible.
In the clinical example above, a patient's future treatment trajectory must account for their disposition toward treatment, mental fortitude, emotional condition, and baseline physiology --- to say nothing of the indefinite range of contingent background events that no model can anticipate in advance.

Under the noun-based paradigm, the only available strategy is to accumulate as many plausibly relevant variables as possible --- from diabetes status and smoking history to clinical notes and multimodal data sources --- in the hope that predictive performance will follow. 
Despite drawing on justified domain knowledge, this strategy is ultimately an act of speculation rather than reasoning inference.
Ironically, in the era of big data and modern AI, achieving strong performance in outcome prediction and survival analysis is no longer the hard problem 
\cite{lee2018deephit, katzman2018deepsurv, huang2023application}.
However, even when the right variables are identified by chance, we remain unable to explain how the effective variables lead to their outcomes --- that is, no account of \emph{causality}. 
These deep learning-based efforts, however sophisticated, ultimately reduce to an exercise in replacing causation with dense correlations.

Researchers with expertise in causal inference have recognized this substitution, and have accordingly emphasized that causal reasoning requires counterfactual modeling \cite{chatha2022dynamic, melnychuk2022causal}.
By partitioning outcomes into factual/counterfactual, classical causal inference appears to offer a structured ``binary choosable future''. Yet it implies a stronger assumption that the variables bearing on the outcome are closed-ended and have been exhaustively specified by the model in advance. This is not speculation but blind assertion \cite{scholkopf2021toward}, functioning as the foundational premise upon which the subsequent inference is dressed as causal reasoning.

\vspace{-4mm}
\begin{figure*}[h]
  \centering
  \includegraphics[width=1.05\textwidth]{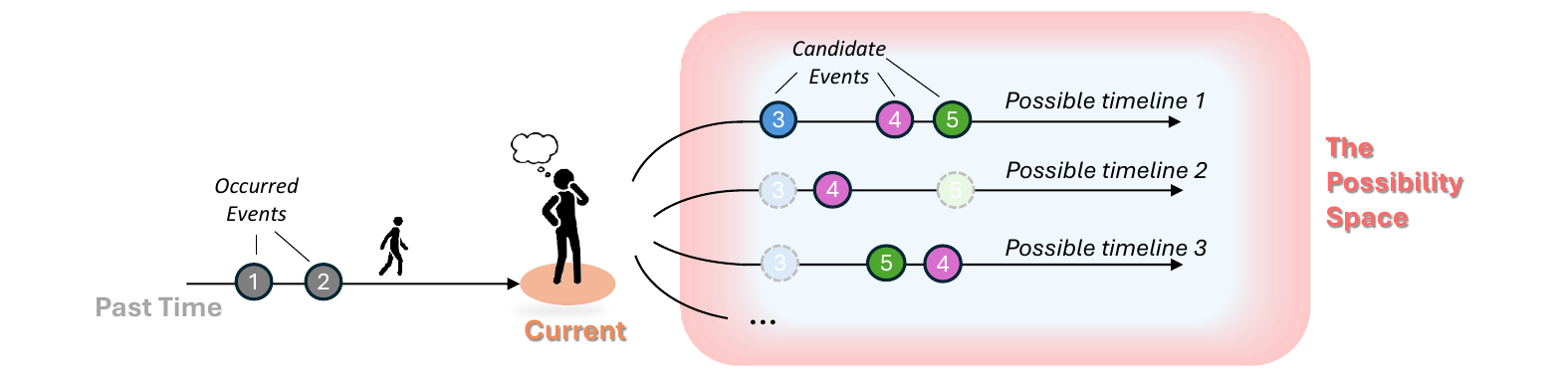}
  \vspace{-3mm}
\caption{The \emph{\textbf{possibility space}} is a branching structure of time-forks, each branch representing an independent possible future that can be named as a \emph{possibility} or a \emph{possible world}. Possibilities generated under identical settings on timing --- or other defining factors --- fall within, and thereby constitute, a \emph{\textbf{hyper-world}} (not reflected in this figure but will be introduced later).}
  \label{fig:forks}
\end{figure*}
\vspace{-2mm}



The future branches into an unbounded number of time-forks (as illustrated in Figure~\ref{fig:forks}), each carrying its own independent dynamics as a possible trajectory of effect. A \emph{choosable future} is therefore defined as a \emph{\textbf{possibility space}} — constituted by all \emph{possibilities} that remain open and worth anticipating. 

The conventional noun-based paradigm, by contrast, collapses the possibility space and fixes a specific future by pre-specifying noun variables as decisive. As shown in Figure~\ref{fig:forks}, an event (e.g., the administration of a particular medicine) may be significant in some possibilities but incidental in others; pre-specifying it as decisive before modeling begins forecloses the very distinction the possibility space is meant to preserve.

Our \emph{causal} thinking naturally operates across possibilities: departing from a targeted future (i.e., the effect) and reasoning backward, we recover the maximal causal efficacy of any contributing factor (i.e., the cause). Armed with this causal knowledge, we can construct a desired future by controlling the relevant causal factors --- thereby enacting a choice through action.
Accordingly, the purpose of causal modeling should be to render such deduction computationally tractable --- furnishing an instrument against which scientific assumptions can be rigorously tested --- rather than adopting those assumptions as pre-validated truths.

In the noun-based paradigm, timing is fixed in advance alongside pre-specified noun variables and need not enter the computation. The verb-based paradigm, by contrast, requires timing to be treated as a computable variable, whose value is not presupposed but derived, to ensure the resulting possibility space is complete.

In classic causal inference, such verb-based thinking has long been recognized, most notably in the do-calculus framework \cite{pearl2012calculus}, where the causal effect of an intervention on input $x$ is expressed as $P(y\mid do(x))$, placing a verb $do(\cdot)$ at the center of causal representation. 
Yet do-calculus ultimately works within the noun-based paradigm rather than replacing it --- in practice, reducing to the identification of conditions under which noun-based models can approximate verb-based causal reasoning \cite{huang2012pearl}.
The required reform runs in the opposite direction: rather than bending verb-based thinking to fit noun-based constraints, the paradigm itself must be rebuilt around verbs --- with timing computation as its necessary foundation.

\section{Timing Learnable $\neq$ Timing Computable}
\vspace{-4mm}

Neural networks trained via backpropagation provide the technical foundation for \emph{verb-based} timing computation --- reasoning backward from the targeted effect to derive the relative timing of contributing causes, rather than presupposing it. This stands in contrast to classical noun-based ``forward'' modeling, in which the functional relationship $f(x)=y$ must be fully specified in advance.
The ``backward'' modeling can be expressed as $x=f^{-1}(y)$, where the relative timing of $x$ is determined post hoc through the modeling process --- with $f^{-1}(y)$ representing the maximal causal efficacy of $x$ with respect to the targeted effect $y$. We formally define this concept in Section~\ref{sec:causal}.

This framework thereby enables causal deduction in a purely data-driven manner. The resulting timing shift for any given event — spanning the possibility space of \emph{earlier}, \emph{later}, or \emph{not at all} — can be evaluated with respect to a fixed target outcome, yielding genuine counterfactual timing deduction without recourse to prior domain knowledge.



Crucially, what a neural network yields is not a deterministic model but \emph{an instantiation of a possibility} --- a single realization drawn from the space of open futures. 
Neural networks exhibit precisely this behavior in practice: even with identical initializations, the learned parameters can differ across training runs --- to say nothing of generative models in which stochastic noise is deliberately introduced as part of the input. In short, while traditional modeling seeks deterministic mappings over noun variables, a neural network enacts a verb.


%


This raises a fundamental question: how should timing computation be formally defined? 
In existing work, many models treat time as a learnable input feature --- from traditional GLMs to contemporary RNN variants. 
Motivated by the irregular sampling inherent to real-world longitudinal data, a growing body of work has emerged to develop temporally adaptive architectures, including: (1) time-aware RNNs that adjust the hidden-state update mechanism within the recurrent cell \cite{baytas2017patient, che2018recurrent, al2024ta}; (2) time representation learning that encodes time as an explicit input embedding \cite{kazemi2019time2vec, shukla2021multi}; and (3) ODE-based methods that model the entire trajectory as continuous dynamics \cite{rubanova2019latent, kidger2020neural}. 
A detailed categorization of related works is provided in the companion paper 
\ifpreprint
\cite{li2026capturetimingattentioneventsclinical}.
\else
\cite{anonymous2026companion}.
\fi

A limitation of these approaches has received insufficient attention: temporal adjustment is designed to align individuals' hidden states over time by assuming their trajectories are globally consistent at the group level --- a constraint well-suited to physical dynamic systems but rarely warranted in clinical data. 
As illustrated in Figure~\ref{fig:forks}, the same set of candidate events (e.g., clinical treatments) may follow different orderings across possible futures, reflecting individual-level non-Markovian dynamics that constitute a principled, largely overlooked basis for patient subtyping.

Where existing methods achieve \emph{timing-learnable} --- the capacity to extract temporal information from observed ground-truth timestamps --- \emph{timing-computable} denotes the strictly broader capability that additionally encompasses \emph{timing-predictable}: the capacity to derive relative timing for candidate events at the individual level, rather than merely observing it.


Notably, virtually no existing work directly addresses timing prediction --- despite the fact that repositioning the input variable $t$ as the outcome in the objective function would seem, at first glance, sufficient to achieve it.
The most critical absence is a formal definition of the \emph{\textbf{possible timing distribution}} — a concept that must be established prior to any methodological discussion. Once a variable is treated as predictable, its predicted values must constitute a distribution rather than a single point estimate; only on this basis can meaningful evaluation be defined. To be fully general, a \emph{possible timing distribution} must hold at two levels: across the cohort for any given event (how does this event's timing distribute over patients?) and across all possibilities for any individual patient (how does this patient's timing distribute over possible futures?). More importantly, such a concept must support reasoning over both outcome $y$ and input $x$.

Under the noun-based paradigm, all entities and events carry fixed, observation-determined timestamps; any estimated timing unconfirmed by ground truth is accordingly treated as inadmissible --- effectively unreal within the paradigm's own terms. 
Timing cannot, by construction, be treated as a computable dimension. 
Enabling timing computation therefore requires the prior concept of a \emph{possibility space} --- a framework within which multiple possible timelines co-exist, each constituting a distinct realization of an open future.

Survival Analysis \cite{wang2019machine} can be understood as an indirect approximation of a possible timing distribution, pursued within the constraints of the noun-based paradigm and without explicit timing computation --- for instance, estimating the probability that patient A survives longer than patient B. 
As the closest existing approach to timing computation, it demonstrates precisely the gap between our practical \emph{grammar of thought} and the current modeling paradigm.


In the possibility space, probability theory remains applicable as is standard in statistical reasoning; what changes, however, is the object to which it is applied. Under the conventional paradigm, the object of inference is a noun variable $X \in \mathbb{R}^n$; under the verb-based paradigm, it becomes a \emph{possibility variable}, denoted $\mathcal{X}=(X,t)\in \mathbb{R}^{n+1}$, with the timing dimension $t$ incorporated as a computable constituent.
Consider the possibility variables for outcome $y$ and input $x$, respectively, we have: 

\begin{definition}[Timing Computation in Neural Network Context] $\\$
\vspace{-5mm}
\begin{itemize}[leftmargin=*]
    \item Let $\mathcal{Y} = (Y, t)$ denote the possibility variable for the target outcome event:
    \begin{itemize}
    \vspace{-1.5mm}
        \item[-] the event's possible timing distribution emerges naturally across the cohort;
        \item[-] an individual's possible timing distribution can be constructed empirically via multiple independent training runs, each yielding a distinct instantiation.
    \end{itemize}
    \item Let $\mathcal{X}=(X, t)$ denote the possibility variable for the causal input event; its \textbf{computable relative timing} $\tau(t)$ is defined within the network architecture via trainable individual-level parameters:
    \begin{itemize}
    \vspace{-1.5mm}
        \item[-] the event's timing distribution is defined relative to the constructed group-level timeline $\tau$;
        \item[-] an individual's timing distribution relative to the specific timeline $\tau$ can be constructed empirically via multiple independent training runs with fixed group-level parameters.
    \end{itemize}
\end{itemize}
\end{definition}

\begin{figure*}[h]
  \hspace{-2mm}
  \includegraphics[width=0.98\textwidth]{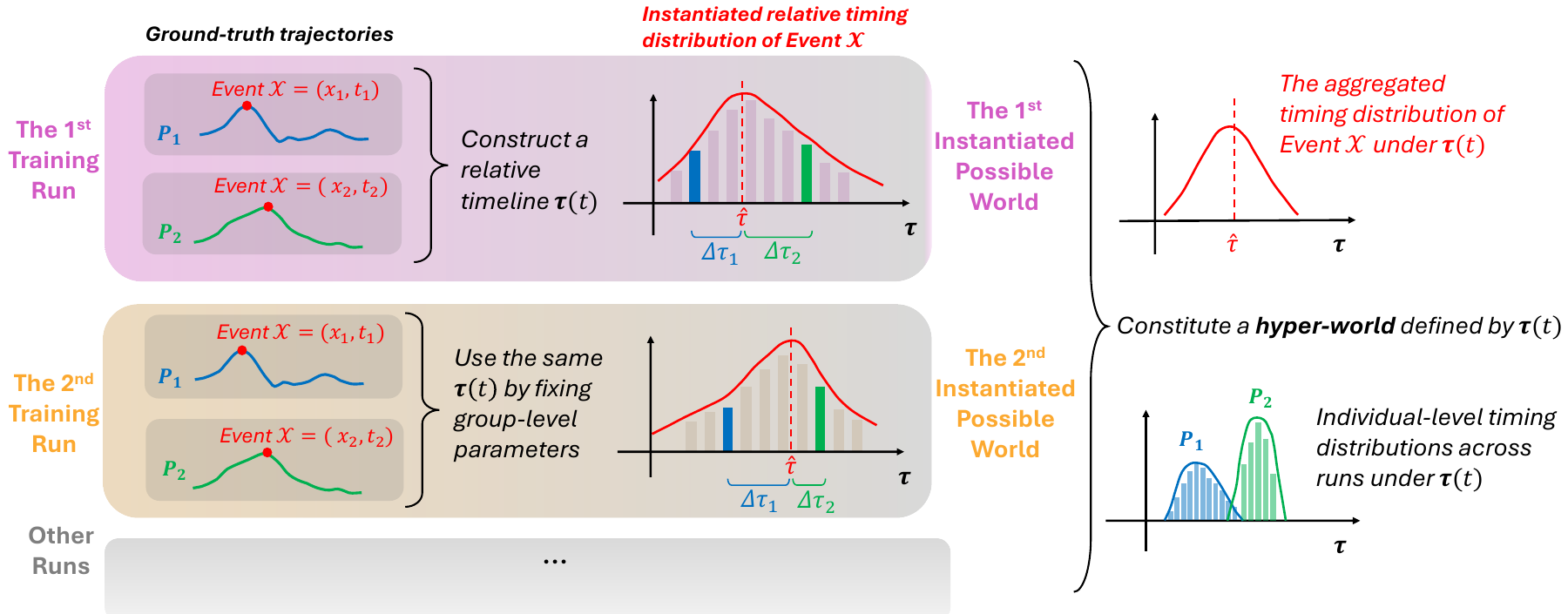}
  \vspace{-2mm}
  \caption{ Independent training runs (instantiating independent possible worlds) under a fixed $\tau(t)$ constitute a \emph{\textbf{hyper-world}}.The individual-level timing distributions for patients $P_1$ and $P_2$ on event $\mathcal{X}$ are built by pooling across runs. }
  \label{fig:runs}
\end{figure*}


Figure \ref{fig:runs} illustrates the timing computation scheme for $\mathcal{X}=(X, t)$ as modeling input. 
Through the first training run, a relative timeline $\tau(t)$ is constructed as a shared group-level reference across all patients, where the individual residuals $\Delta \tau_1$ and $\Delta \tau_2$ capture each patient's temporal characteristics. 
By fixing the group-level parameters and performing multiple independent runs, multiple \emph{possible worlds} are instantiated under the same $\tau(t)$ --- constituting a \emph{hyper-world} defined by $\tau(t)$. 
Within this hyper-world, the possible timings of different patients within the same run, and of the same patient across runs, are guaranteed to remain \emph{comparable}. 
As an unbounded number of possible worlds can be instantiated under $\tau(t)$, the hyper-world enables an unbounded capacity for virtual trial and error.
Such comparability is the technical foundation of counterfactual deduction: by reassigning the target outcome to construct a counterfactual effect, the derived shift in the individual-level timing distribution naturally yields a What-If answer.

Temporal relationships among multiple input events can be investigated across different $\tau(t)$ instantiations --- that is, across different hyper-worlds. For example, if 100 independent $\tau(t)$ constructions are performed, and approximately 90 exhibit event $\mathcal{X}_a$ preceding event $\mathcal{X}_b$ while 10 show the reverse order, the latter may be identified as a rare ($10\%$) but potentially significant temporal pattern --- particularly if this ordering shift corresponds to meaningful outcome changes. 
In clinical practice, such patterns could provide a principled basis for patient subtyping. 
More broadly, the temporal relationships among individual patients can empirically recover survival-analysis-style conclusions — grounded in an explicit possibility space rather than indirect approximation.

We term this overall scheme \emph{\textbf{possibility emergence}}, grounded in the backpropagation mechanism of neural networks, where ``emergence'' is a term borrowed from \emph{Causal Emergence} in complex systems \cite{yuan2024emergence}. Metaphorically, this is to use AI as the \emph{dice of possibilities}. For the technical implementation of group- and individual-level parameters, we refer readers to the companion paper 
\ifpreprint
\cite{li2026capturetimingattentioneventsclinical}.
\else
\cite{anonymous2026companion}.
\fi

\section{Applications of Possible Timing Distribution}

We instantiate the proposed paradigm on real-world longitudinal EHR data from 3,276 breast cancer patients, predicting the onset timing of cardiotoxicity-induced heart disease. We define the \emph{\textbf{timing attention}} of an event as the degree of concentration of its timing distribution across the patient cohort, measured via kurtosis.

The demonstrated applications include: (1) the automatic discovery of clinically significant patient trajectories, and (2) the deduction of counterfactual event timing for positive patients (as if they remain negative). 
All results are model-derived in a purely data-driven manner — to our knowledge, the first such demonstration in the machine learning literature for both tasks.
In the following, we illustrate the effectiveness of these two applications through representative examples; interactive scripts for independent exploration are made available online
\footnote{\url{https://github.com/kflijia/LITT_python_code.git}}

\vspace{-2mm}
\begin{figure*}[h]
  \centering
  \includegraphics[width=0.99\textwidth]{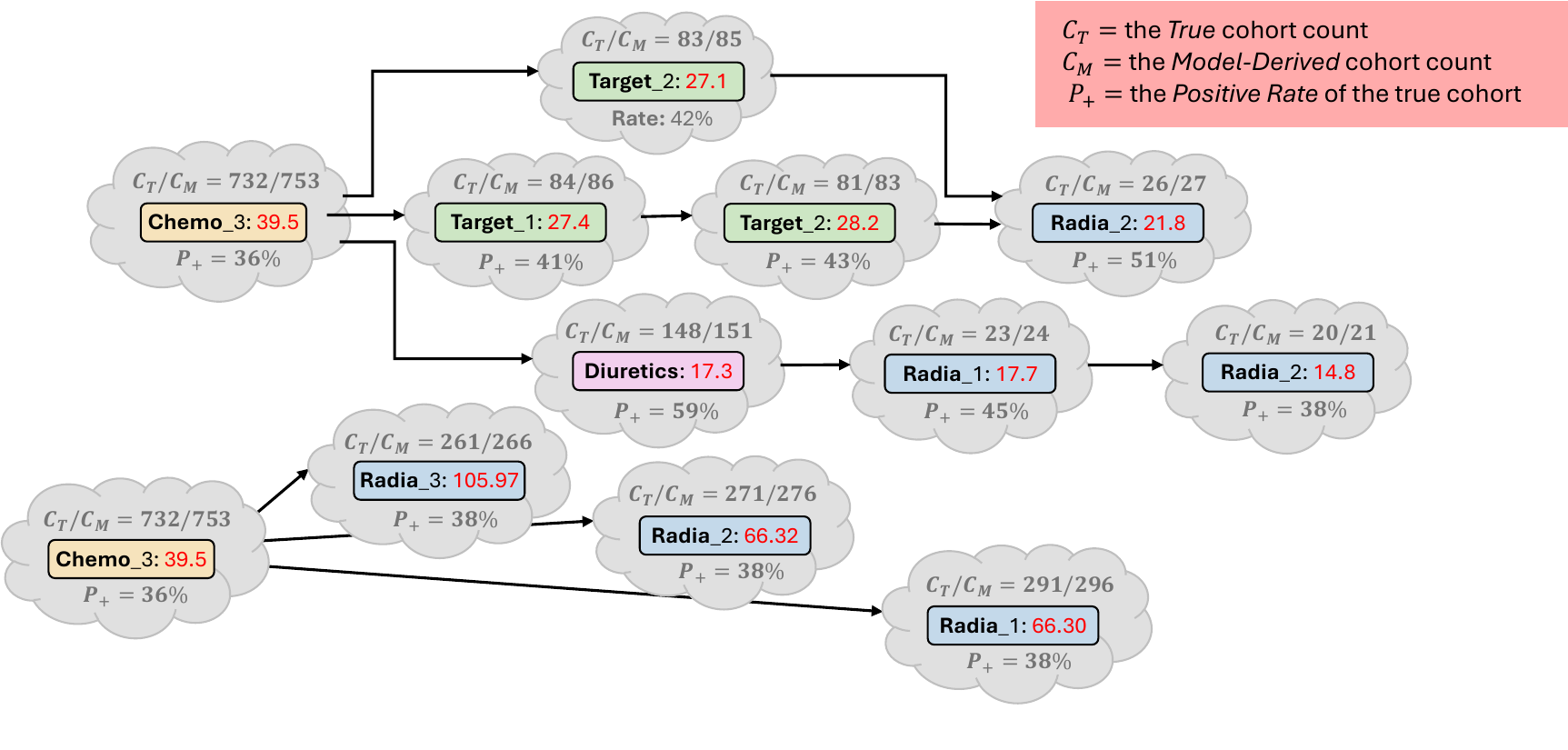}
  \vspace{-4mm}
  \caption{Two representative trajectory clusters automatically discovered by the model. Red values denote the \emph{timing attention} $\kappa$ of each event; $C_M$ is the count of model-derived core candidate patients (after excluding kurtosis outliers), and $C_T$ is the observed true count within $C_M$. Numeric suffixes denote administration order (e.g., $\text{Chemo}\_3 =$ third chemotherapy).}
  \label{fig:exp1_tro}
\end{figure*}

Figure \ref{fig:exp1_tro} presents remarkably interpretable trajectories. Three observations reflect the model's effectiveness: events with larger cohort sizes are consistently associated with higher timing-attention $\kappa$ values; $C_T$ remains close to $C_M$ throughout each trajectory; and cohort positive rates exhibit clear trends across successive steps.

In the top cluster, the first two pathways merge at the shared terminal event $\text{Radia}\_2$ (the second radiation therapy), indicating that their intermediate segments carry equivalent temporal significance — the significance of $\text{Target}\_1$ (the first targeted therapy) is fully subsumed by $\text{Target}\_2$, which emerges as the decisive event along both pathways.
The third pathway also terminates at $\text{Radia}\_2$ but remains distinct, as reflected in its differing $\kappa, C_T, C_M$, and $P_{+}$ values. Here, diuretics, medications that support kidney function and manage heart failure, act as a critical turning point: $P_{+}$ peaks at $59\%$ at this step before declining progressively, in clear contrast to the monotonically increasing trend observed in the other two pathways.

The bottom cluster highlights a key methodological advantage: The timing-attention mechanism evaluates events concurrently, without imposing logical ordering constraints, avoiding the exponential search complexity of traditional association rule-based approaches. Notably, all three terminating events share $P_{+}=38\%$ despite increasing $\kappa$ values, reflecting that most patients in this cohort ultimately undergo three rounds of radiation, not implying that additional radiation administrations are more predictive of cardiotoxicity.

\vspace{-1mm}
\begin{figure*}[h]
  \centering
  \includegraphics[width=0.99\textwidth]{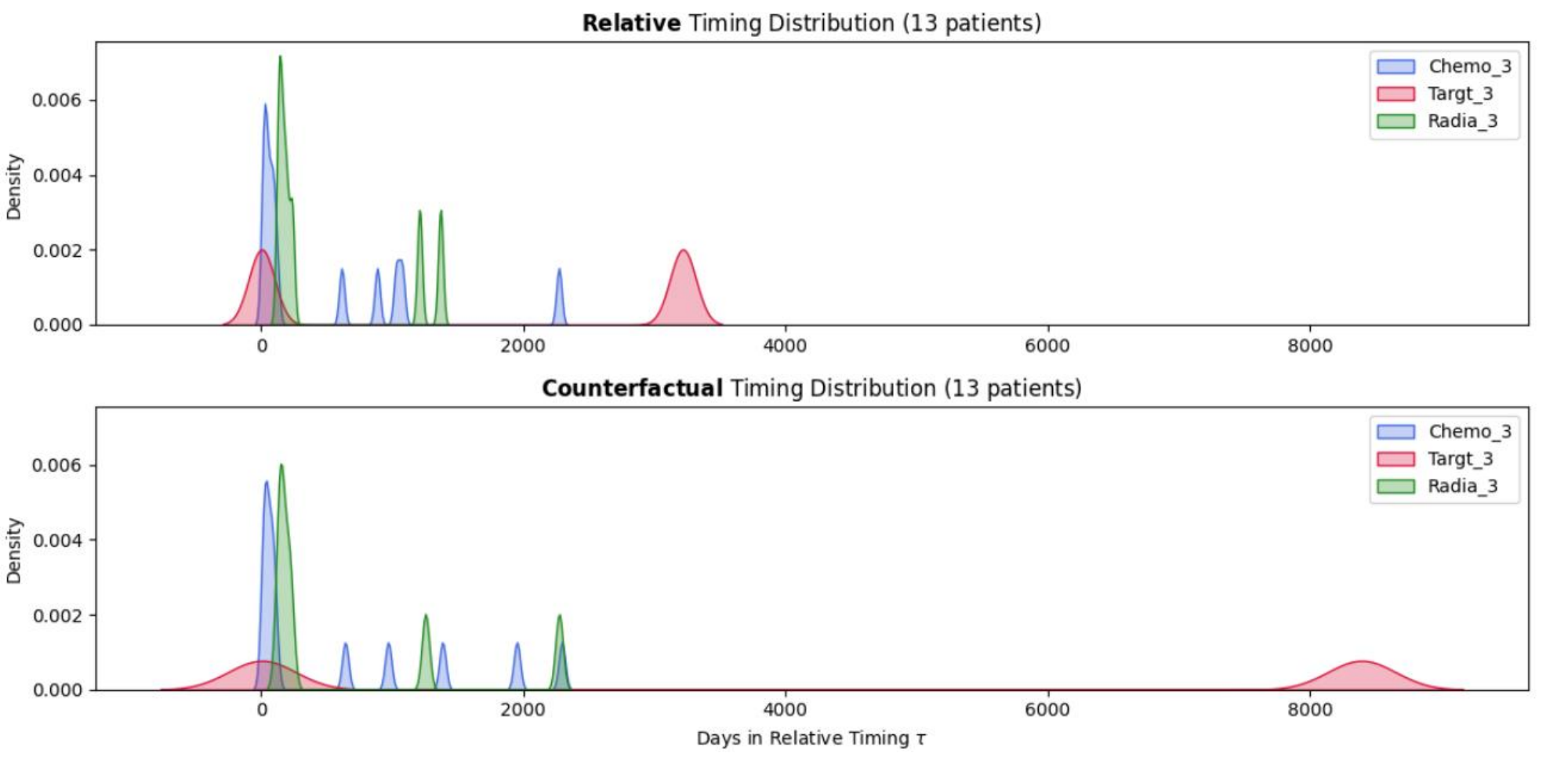}
  \vspace{-3mm}
  \caption{KDE plots of model-derived relative and counterfactual timing distributions for three events ($\text{Chemo}\_3$, $\text{Target}\_3$, and $\text{Radia}\_3$), under the counterfactual assumption that positive patients following the trajectory $\text{Dx\_DB}\_2 \rightarrow \text{Dx\_HT}\_3 \rightarrow \text{Chemo}\_3$ are negative with respect to cardiotoxicity. $\text{Dx\_DB}$ and $\text{Dx\_HT}$ denote diagnoses of diabetes and hypertension, respectively.}
  \label{fig:exp2_flip}
\end{figure*}

Figure \ref{fig:exp2_flip} compares the model-derived counterfactual and original relative timing distributions of the selected cohort within the shared timeline $\tau$. We observe that for approximately half of the cohort, the relative timing of $\text{Target}\_3$ shifts markedly toward larger values under the counterfactual assumption. 
This suggests that postponing (or canceling, where the counterfactual timing exceeds the patient's relative observation window in $\tau$) the third targeted therapy could substantially reduce the risk of cardiotoxicity onset for these patients.

\section{Toward Causal-Reasoning World Model} \label{sec:causal}


In conventional causal inference, a variable's factual and counterfactual values are pre-specified as a binary encoding (e.g.,$y=1/0$) before any modeling begins.
However, the intended causal efficacy carries far richer conceptual content than any such binary can capture. 
In cancer care, for instance, survival and death are the conventional proxies for treatment success, yet a gradually declining quality of life between them may be clinically more relevant.

This reveals a fundamental limitation: we cannot verify whether the specified variable adequately represents the causal relationship we aim to model, and empirically, it rarely does. 
Within the verb-based paradigm, by contrast, causal events can be empirically represented through possibility emergence, as determined through backpropagation from the targeted effect.


\begin{definition}[Causal Factum] $\\$
\vspace{-1mm}
 $\\$
A timing-computable possibility variable $\mathcal{X}=(X,\tau(t))$ ranges over emergent possibilities whose \textbf{domain} is determined through backpropagation from the targeted effect $\mathcal{Y}$ --- thereby defining a \textbf{causal factum} $\mathcal{X}_{emg(\mathcal{Y})}$, representing the maximal causal efficacy of $\mathcal{X}$ with respect to $\mathcal{Y}$.
\end{definition}


Figure~\ref{fig:exp2_flip} provides a direct illustration: the ground-truth input $\mathcal{X} = (X, t)$ contains the full treatment history of these patients across three activity categories --- chemotherapy, targeted therapy, and radiation therapy --- while $Y = 1/0$, which indicates the diagnosis status of cardiotoxicity-induced heart disease, serves as the targeted effect. Since the two displayed timing distributions share the same $\tau(t)$, they constitute two comparable instantiations of the possibility variable $\mathcal{X} = (X, \tau(t))$.
As $\text{Target}\_3$ appears to be the only event with a significant timing shift among all candidate events in $\mathcal{X}$, the causal relationship can be expressed as $\mathcal{X}_{emg(Y)} \rightarrow Y$, that is, $\text{Target}\_3 \rightarrow Y$ in this case.

Notably, within the verb-based paradigm, the ``noun cause'' $X$ (i.e., $\text{Target}\_3$ in the example) is determined post hoc from the constructed hyper-world through possibility emergence --- hence the term causal \emph{\textbf{deduction}}, in contrast to traditional causal \emph{inference}. The former is amenable to realization as a purely data-driven procedure, operated automatically by AI; the latter, by contrast, produces inference only about a manually pre-specified $X$.

The noun-based causal inference framework is structurally prone to confounding: Because the pre-specified variable $X$ cannot, in general, perfectly match the cause required by the objective relationship $X \rightarrow Y$, the resulting information residual $I(X - X_{emg(Y)})$ must therefore be attributed to an additional noun variable --- which is, a confounder. 
However, since no observed variable can perfectly represent the required confounder, its specification introduces a new residual, which in turn demands yet another confounder. 
This is not an incidental modeling failure but a structural regress inherent to any framework that requires causes to be pre-specified as noun variables. 

It is worth noting that this confounding problem presupposes a closed, pre-specified causal graph --- a noun-based assumption that the verb-based paradigm dissolves by construction. Within this paradigm, causal structure is not assumed in advance but emerges spontaneously through possibility modeling; confounders do not arise because no noun variable is pre-specified to mismatch against.

More precisely, the required causal factum $\mathcal{X}_{emg(Y)}$ can be fully represented through emergent possibilities, without requiring additional variables to account for unexplained variance. More importantly, the explicit effect residual $I(\mathcal{Y} - \mathcal{X}_{emg(Y)})$ naturally motivates AI to extend the current causal structure spontaneously --- pursuing the upstream cause that accounts for this residual, and in doing so, incrementally constructing an ever-richer causal reasoning graph. This is the foundation of a \emph{spontaneously growing causal-reasoning world model}: one that is not pre-engineered but emerges through the verb-based process of reasoning from effects toward causes.

Today, AI continues to build underlying representations increasingly disconnected from human understanding. LLMs, for instance, operate within a latent space of statistical token associations, without grounding in human conceptual or linguistic structure. There is an urgent need to enable a new class of generative models \footnote{\url{https://radical.vc/building-spatially-intelligent-ai/}} --- referred to as \emph{emergence models} in the verb-based paradigm --- capable of empirically constructing hyper-worlds that are structurally aligned with human thought and extend beyond the observed real world.


\section*{Acknowledgement}
\ifpreprint
\vspace{-2mm}
\fi
The philosophical framework underlying this work draws deeply on the thought of Prof. Zhao Tingyang, whose writings have served as an intellectual beacon throughout.
This research received no external funding.





\bibliography{main}

@article{anonymous2026companion,
  title={Capture Timing Attention in Clinical Time Series},
  author={Anonymous},
  journal={Transactions on Machine Learning Research},
  year={2026},
  note={Under review. \url{https://openreview.net/forum?id=ym8VCQLgHZ}}
}

@misc{li2026capturetimingattentioneventsclinical,
      title={Capture Timing-Attention of Events in Clinical Time Series}, 
      author={Jia Li and Yu Hou and Rui Zhang},
      year={2026},
      eprint={2602.10385},
      archivePrefix={arXiv},
      primaryClass={cs.LG},
      url={https://arxiv.org/abs/2602.10385}, 
}

@inproceedings{baytas2017patient,
  title={Patient subtyping via time-aware LSTM networks},
  author={Baytas, Inci M and Xiao, Cao and Zhang, Xi and Wang, Fei and Jain, Anil K and Zhou, Jiayu},
  booktitle={Proceedings of the 23rd ACM SIGKDD international conference on knowledge discovery and data mining},
  pages={65--74},
  year={2017}
}

@article{al2024ta,
  title={TA-RNN: an attention-based time-aware recurrent neural network architecture for electronic health records},
  author={Al Olaimat, Mohammad and Bozdag, Serdar and Alzheimer’s Disease Neuroimaging Initiative},
  journal={Bioinformatics},
  volume={40},
  number={Supplement\_1},
  pages={i169--i179},
  year={2024},
  publisher={Oxford University Press}
}

@article{kazemi2019time2vec,
  title={Time2vec: Learning a vector representation of time},
  author={Kazemi, Seyed Mehran and Goel, Rishab and Eghbali, Sepehr and Ramanan, Janahan and Sahota, Jaspreet and Thakur, Sanjay and Wu, Stella and Smyth, Cathal and Poupart, Pascal and Brubaker, Marcus},
  journal={arXiv preprint arXiv:1907.05321},
  year={2019}
}

@article{shukla2021multi,
  title={Multi-time attention networks for irregularly sampled time series},
  author={Shukla, Satya Narayan and Marlin, Benjamin M},
  journal={arXiv preprint arXiv:2101.10318},
  year={2021}
}

@article{rubanova2019latent,
  title={Latent ordinary differential equations for irregularly-sampled time series},
  author={Rubanova, Yulia and Chen, Ricky TQ and Duvenaud, David K},
  journal={Advances in neural information processing systems},
  volume={32},
  year={2019}
}

@article{kidger2020neural,
  title={Neural controlled differential equations for irregular time series},
  author={Kidger, Patrick and Morrill, James and Foster, James and Lyons, Terry},
  journal={Advances in neural information processing systems},
  volume={33},
  pages={6696--6707},
  year={2020}
}

@inproceedings{lee2018deephit,
  title={Deephit: A deep learning approach to survival analysis with competing risks},
  author={Lee, Changhee and Zame, William and Yoon, Jinsung and Van Der Schaar, Mihaela},
  booktitle={Proceedings of the AAAI conference on artificial intelligence},
  volume={32},
  number={1},
  year={2018}
}

@article{katzman2018deepsurv,
  title={DeepSurv: personalized treatment recommender system using a Cox proportional hazards deep neural network},
  author={Katzman, Jared L and Shaham, Uri and Cloninger, Alexander and Bates, Jonathan and Jiang, Tingting and Kluger, Yuval},
  journal={BMC medical research methodology},
  volume={18},
  number={1},
  pages={24},
  year={2018},
  publisher={Springer}
}

@inproceedings{melnychuk2022causal,
  title={Causal transformer for estimating counterfactual outcomes},
  author={Melnychuk, Valentyn and Frauen, Dennis and Feuerriegel, Stefan},
  booktitle={International conference on machine learning},
  pages={15293--15329},
  year={2022},
  organization={PMLR}
}

@article{chatha2022dynamic,
  title={Dynamic survival transformers for causal inference with electronic health records},
  author={Chatha, Prayag and Wang, Yixin and Wu, Zhenke and Regier, Jeffrey},
  journal={arXiv preprint arXiv:2210.15417},
  year={2022}
}

@article{wang2019machine,
  title={Machine learning for survival analysis: A survey},
  author={Wang, Ping and Li, Yan and Reddy, Chandan K},
  journal={ACM Computing Surveys (CSUR)},
  volume={51},
  number={6},
  pages={1--36},
  year={2019},
  publisher={ACM New York, NY, USA}
}

@article{huang2023application,
  title={Application of machine learning in predicting survival outcomes involving real-world data: a scoping review},
  author={Huang, Yinan and Li, Jieni and Li, Mai and Aparasu, Rajender R},
  journal={BMC medical research methodology},
  volume={23},
  number={1},
  pages={268},
  year={2023},
  publisher={Springer}
}

@article{yuan2024emergence,
  title={Emergence and causality in complex systems: a survey of causal emergence and related quantitative studies},
  author={Yuan, Bing and Zhang, Jiang and Lyu, Aobo and Wu, Jiayun and Wang, Zhipeng and Yang, Mingzhe and Liu, Kaiwei and Mou, Muyun and Cui, Peng},
  journal={Entropy},
  volume={26},
  number={2},
  pages={108},
  year={2024},
  publisher={MDPI}
}

@article{che2018recurrent,
  title={Recurrent neural networks for multivariate time series with missing values},
  author={Che, Zhengping and Purushotham, Sanjay and Cho, Kyunghyun and Sontag, David and Liu, Yan},
  journal={Scientific reports},
  volume={8},
  number={1},
  pages={6085},
  year={2018},
  publisher={Nature Publishing Group UK London}
}

@article{pearl2012calculus,
  title={The do-calculus revisited},
  author={Pearl, Judea},
  journal={arXiv preprint arXiv:1210.4852},
  year={2012}
}

@article{scholkopf2021toward,
  title={Toward causal representation learning},
  author={Sch{\"o}lkopf, Bernhard and Locatello, Francesco and Bauer, Stefan and Ke, Nan Rosemary and Kalchbrenner, Nal and Goyal, Anirudh and Bengio, Yoshua},
  journal={IEEE},
  volume={109},
  number={5},
  pages={612--634},
  year={2021},
  publisher={IEEE}
}

@article{huang2012pearl,
  title={Pearl's calculus of intervention is complete},
  author={Huang, Marco Valtorta, Yimin },
  journal={ arXiv:1206.6831},
  year={2012}
}
\bibliographystyle{tmlr}











\end{document}